\pdfoutput=1

\documentclass[11pt]{article}

\usepackage{EMNLP2023}

\usepackage{times}
\usepackage{latexsym}

\usepackage[T1]{fontenc}

\usepackage[utf8]{inputenc}

\usepackage{microtype}

\usepackage{inconsolata}

\usepackage{amsmath}
\usepackage{amssymb}
\usepackage{array}
\usepackage{lipsum}

\usepackage{multirow}
\usepackage{caption}
\usepackage{subcaption}
\usepackage{booktabs}
\usepackage{soul} 
\usepackage{xcolor}
\usepackage{graphicx}
\usepackage{longtable}

\title{Knowledge Corpus Error in Question Answering}

\author{Yejoon Lee\thanks{\hspace{.2cm}Work done during internship at KAIST AI} \\ Seoul National University\\ \url{leeyejoon@snu.ac.kr}
\And Philhoon Oh\\ KAIST AI \\ \url{philhoonoh@kaist.ac.kr}
\And James Thorne \\ KAIST AI\\ \url{thorne@kaist.ac.kr}}

\begin{document}
\maketitle
\begin{abstract}
Recent works in open-domain question answering (QA) have explored generating context passages from large language models (LLMs), replacing the traditional retrieval step in the QA pipeline. However, it is not well understood why generated passages can be more effective than retrieved ones. This study revisits the conventional formulation of QA and introduces the concept of \textit{knowledge corpus error}. This error arises when the knowledge corpus used for retrieval is only a subset of the entire string space, potentially excluding more helpful passages that exist outside the corpus. LLMs may mitigate this shortcoming by generating passages in a larger space. We come up with an experiment of paraphrasing human-annotated gold context using LLMs to observe knowledge corpus error empirically. Our results across three QA benchmarks reveal an increased performance (10\% - 13\%) when using paraphrased passage, indicating a signal for the existence of knowledge corpus error.\footnote{Our code is available at \url{https://github.com/xfactlab/emnlp2023-knowledge-corpus-error}} 
\end{abstract}

\section{Introduction}

Large language models (LLMs) generate surprisingly fluent and informative texts. This led to many works utilizing the text data generated by these models for purposes such as instruction tuning \citep{honovich2022unnatural, wang_self-instruct_2023} and improving reasoning capability \citep{zelikman_star_2022, ho2023large, magister2023teaching}.

\begin{figure}[!t]
  \centering
  \includegraphics[width=1\linewidth]{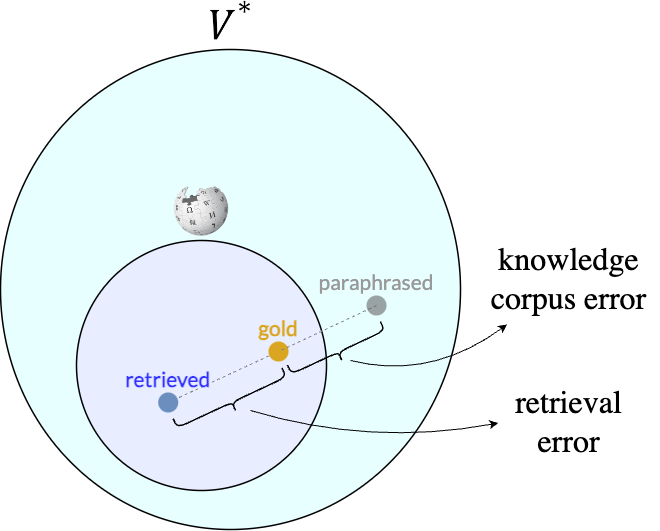}
  \caption{Illustration of our framework. Each dot represents a passage. \textit{Retrieval error} refers to an error from failing to retrieve the gold passage. \textit{Knowledge corpus error} refers to an error from discarding better passages outside the knowledge corpus, e.g., Wikipedia, which is inevitable in any retrieval setting.
  See §\ref{body:analytic} for details.}
  \label{fig:intro}
\end{figure}

Open-domain question answering (QA) \citep{chen-etal-2017-reading} is a task where retrieving relevant passages from a corpus of factual information such as Wikipedia is standard practice.
Recent works have attempted to generate such passages from LLMs, replacing the retrieval step of the traditional pipeline \citep{sun2023recitationaugmented, yu2023generate}. Despite their success, it is not well understood why these generated passages could be more effective than retrieved passages. These recent advancements lack robust links to prior research in QA posing a challenge to a holistic understanding.

By revisiting the formulation of answer generation with retrieved passages \citep{guu_retrieval_2020, lewis2020rag, singh_end--end_2021}, we identify the then-overlooked gap, which has become significant now due to the advance in LLMs \citep{brown_language_2020}. Our discussion starts with the observation that the \textit{knowledge corpus} from which the passages are retrieved is only a subset of the possible string space. More helpful passages to the reader may exist outside the knowledge corpus. Unfortunately, retrieval, by definition, cannot utilize passages outside the knowledge corpus, potentially causing a shortfall. We refer to this as \textit{knowledge corpus error}. In contrast, LLMs can generate passages from the entire string space, which may mitigate the inherent limits of retrieval.

We empirically demonstrate the presence of knowledge corpus error where a passage from outside of Wikipedia outperforms the human-annotated gold passage inside Wikipedia in question answerin. We design an experiment of paraphrasing human-annotated gold context with LLMs. Experiments with four QA benchmarks, NQ \citep{kwiatkowski-etal-2019-natural}, HotPotQA \citep{yang-etal-2018-hotpotqa}, StrategyQA \citep{geva-etal-2021-aristotle}, and QASC \citep{Khot2019QASCAD} result in 10\% - 13\% gain in reader performance across three benchmarks, when using paraphrased passages. This gain supports our hypothesis that there exist more helpful passages than the gold passage outside the knowledge corpus.

\section{Related Work}

\subsection{Leveraging LLM-generated text}
As the quality of text generated from LLMs has improved through larger models \citep{kaplan2020scaling} and instruction tuning \citep{sanh2022multitask, wei2022finetuned}, many works have sought to use these models as data sources in other NLP tasks. Text generated from LLMs has been used for generating datasets for instruction finetuning \citep{honovich2022unnatural, wang_self-instruct_2023}, improving reasoning \citep{zelikman_star_2022, ho2023large, magister2023teaching}, and many other purposes \citep{liu-etal-2022-wanli, ye-etal-2022-zerogen, haluptzok2023language}.

Recently, there has been growing attention towards open-source LLMs \citep{touvron2023llama} finetuned on instructions generated from proprietary LLMs, such as Alpaca \citep{alpaca}, Koala \citep{koala_blogpost_2023} and Vicuna \citep{vicuna2023}. Text generated from these models purportedly match quality of those from proprietary LLMs \citep{vicuna2023}, but this assertion remains disputed \citep{gudibande2023false}. Understanding the role of LLM-generated text will serve as an important aspect of this discourse.

\subsection{Knowledge-intensive NLP and retrieval}
Knowledge-intensive NLP, such as open-domain QA \citep{chen-etal-2017-reading} and fact verification \citep{thorne-etal-2018-fever}, requires substantial factual knowledge that may change over time. Therefore, these tasks were originally envisioned to incorporate the retrieval of relevant passages from the knowledge corpus \citep{chen-etal-2017-reading}. In the typical retrieve-then-read pipeline (\citealt{karpukhin-etal-2020-dense}, \textit{inter alia}), a pipeline of models, first selects \textit{k} passages from a retrieval function which are then used to condition answer generation from reader (\citealt{izacard-grave-2021-leveraging}, \textit{inter alia}).

Meanwhile, the success of pre-trained language models \citep{2020t5} and the associative memory properties learned during training \citep{petroni-etal-2019-language} has allowed researchers to revisit closed-book QA, in which models answer the questions without being provided a passage. Closed-book QA was demonstrated to be effective both in in-context learning \citep{brown_language_2020} and supervised learning \citep{roberts-etal-2020-much}. 

Recent works on chain-of-thought prompting has shown that generating intermediate steps before giving an answer improves the reasoning capability of LLMs \citep{wei_chain--thought_2022, wang2023selfconsistency}. Inspired by this, recent works prompt LLMs to generate the intermediate step of QA, which is the passages \citep{sun2023recitationaugmented, yu2023generate}. These passages are subsequently fed into the reader, either supervised FiD \citep{izacard-grave-2021-leveraging} or in-context learning LLMs \citep{brown_language_2020}.
Despite their success, these methods require a very large scale and risk generated passages containing stale or non-factual information. Moreover, it is not fully explained why generating passages may have advantages over retrieval.

\section{Analytic Discussion}
\label{body:analytic}
Our task formulation follows retrieval augmented models for QA \citep{guu_retrieval_2020, lewis2020rag, singh_end--end_2021}. These works view contexts as a latent variable for the QA model \citep{lee-etal-2019-latent}. 
\subsection{Setup}
Let \(V^*\) be the infinite set of all possible strings over vocabulary tokens in $V$, including the empty string. 
An instance of a QA dataset consists of a triple \((q,a,c)\):
question \(q\), answer \(a\), and context \(c\), where \(q, a, c \in V^*\). Typically, the context \(c\) is retrieved from the knowledge corpus \(\mathcal{Z}\), such as Wikipedia, where \(\mathcal{Z} \subset V^*\).

\subsection{QA Task Formulation}
The goal of QA is to learn a distribution \(p(a|q)\), where models decode a string $a$ that acts as an abstractive answer to the query \citep{lewis2020rag, izacard-grave-2021-leveraging}.
One can directly prompt a language model to obtain an answer \(a\), given question \(q\) (where context $c$ is implicitly the empty string), relying only on model parameters in closed-book QA \citep{roberts-etal-2020-much, brown_language_2020}.

\begin{equation}
\hat{a} = \operatorname{arg\,max}_{a \in V^*} \space p(a|q)
\end{equation}

However, direct prompting is often difficult to learn and barely discloses its inner working. Therefore, a popular approach is to marginalize \(p(a|q)\) over contexts in the knowledge corpus \citep{guu_retrieval_2020, lewis2020rag, singh_end--end_2021}.
As it is intractable to calculate the probability for all the contexts in the knowledge corpus, \(p(a|q)\) is approximated to the sum of probabilities for top \textit{k} contexts from \(\mathcal{Z}\). \(Topk(\mathcal{Z}, q)\) denotes the set of resulting top \textit{k} passages after the retrieval with a query \(q\).

\begin{equation}
    p(a|q) \approx \sum_{c\in Topk(\mathcal{Z}, q)}p(a|q,c)p(c|q)
\end{equation}

The gap in this formulation is that relevant context \(c\) may exist outside of the knowledge corpus \(\mathcal{Z}\). This makes the sum of marginal probabilities over \(\mathcal{Z}\) only an approximation. The true probability would require the summation of marginal probabilities over the infinite string space \(V^*\).

\begin{equation}
\begin{split}
    p(a|q) &{=} \sum_{c\in S} p(a|q,c)p(c|q) \\
    &{\approx} \sum_{c\in \mathcal{Z}}p(a|q,c)p(c|q) \\
    &{\approx} \sum_{c\in Topk(\mathcal{Z}, q)}p(a|q,c)p(c|q)
\end{split}
\label{eq:eq}
\end{equation}

\subsection{Knowledge corpus error}
Equation~\ref{eq:eq} details two steps of approximation, which results in two sources of potential error in QA using contexts.
The first source of error is introduced when the entire knowledge corpus \(\mathcal{Z}\) is approximated to top \textit{k} retrieved contexts, \(Topk(\mathcal{Z},q)\). This error, which we denote \textit{\textbf{retrieval error}}, can be mitigated by better retrieval methods or increasing \textit{k}, the number of contexts.
On the other hand, the second source of error is introduced when the entire string space \(V^*\) is approximated to knowledge corpus \(\mathcal{Z}\). This error is rooted in the use of knowledge corpus itself, hence we denote it as \textit{\textbf{knowledge corpus error}}. To elaborate, for some \(\tilde{c} \notin \mathcal{Z}, p(\tilde{c}|q) > p(c \in \mathcal{Z}|q)\), but \(p_{retriver}(\tilde{c}|q)=0\) whereas \(p_{LLM}(\tilde{c}|q) > 0\).

For a query \(q\), \(p(c|q)\) is sufficiently small for most contexts \(c\). This allows these terms be ignored by setting \(p(c|q)\) to zero. For instance, top-\textit{k} retrieval is essentially setting \(p(c|q)\) to zero for \(c \notin Topk(\mathcal{Z}, q)\). For contexts outside the knowledge corpus, \(\tilde{c} \notin \mathcal{Z}\),  
applying Bayes' rule, \(p(\tilde{c}|q) \propto p(q|\tilde{c})p(\tilde{c})\), where the retrieval-based task formulation is setting the prior \(p(\tilde{c}) =0 \). 
Knowledge corpus error may explain why reader models can benefit from generated contexts \citep{sun2023recitationaugmented, yu2023generate} as LLMs can generate strings from the set $V^* \supset \mathcal{Z}$.

\begin{table*}[t]
\centering
\begin{tabular}{c||cccccc|c}
\hline
Benchmarks & \multicolumn{3}{c}{Reader: GPT} & \multicolumn{3}{c|}{Reader: Claude} & \small{Average gap between} \\
 & Gold & GPT & Claude & Gold & GPT & Claude & \small{gold and paraphrased} \\
\hline
NQ \small{exact match (\%)} & 40.9 & \textcolor{blue}{39.9} & \textcolor{red}{44.3} & 18.3 & \textcolor{red}{21.3} & \textcolor{red}{35.5} & \textcolor{red}{3.125} \\
HotPotQA \small{exact match (\%)} & 36.3 & \textcolor{red}{38.6} & \textcolor{red}{43.4} & 47.6 & \textcolor{red}{50.9} & \textcolor{red}{54.2} & \textcolor{red}{4.825} \\
StrategyQA \small{accuracy (\%)} & 54.6 & \textcolor{red}{56.4} & \textcolor{red}{70.5} & 68.9 & \textcolor{red}{75.5} & \textcolor{red}{76.5} & \textcolor{red}{7.975} \\
QASC \small{accuracy (\%)} & 95.7 & \textcolor{blue}{92.4} & \textcolor{blue}{91.1} & 86.3 & \textcolor{blue}{75.7} & \textcolor{blue}{76.9} & \textcolor{blue}{- 6.975}\\
\hline
\end{tabular}
\caption{Performance of each reader when given original gold context ("Gold"), paraphrased context with GPT ("GPT"), and paraphrased context with Claude ("Claude"). \textcolor{red}{Red} indicates an \textcolor{red}{increase} in performance after paraphrasing, implying knowledge corpus error has been observed. \textcolor{blue}{Blue} indicates a \textcolor{blue}{decrease} in performance after paraphrasing, implying knowledge corpus error has not been observed.}
\label{tab:mainexp}
\end{table*}

\section{Empirical Observation}
To observe knowledge corpus error, we study the effect of \textit{paraphrasing} human-annotated \textit{gold} contexts from QA dataset. Gold context \(c_{gold} \in \mathcal{Z}\) is what humans annotated as the supporting passage for a given question-answer pair. While human annotation may be imperfect, we assume that this \(c_{gold}\) acts as the best available passage from the knowledge corpus \(\mathcal{Z}\), i.e., there is no retrieval error. In our experiment, \(c_{gold}\) is paraphrased into \(c_{paraph}\), by prompting LLMs with \(c_{gold}\) and \(q\). Then, \(c_{gold}\) and \(c_{paraph}\) are separately fed into the reader to compare the performance.
As \(c_{gold}\) is the best available context, any gains from paraphrasing should be attributed to reduced knowledge corpus error.

\subsection{Experimental setup}
For a single instance of a QA dataset \((q,c_{gold},a)\) and a paraphrased context \(c_{paraph} = Paraph(c_{gold}, q)\), we compare model performance in two settings without and with paraphrasing: \(Read(q, c_{gold})\) and \(Read(q, c_{paraph})\). Both \(Paraph()\) and \(Read()\) are function calls to LLMs, GPT-3.5 (gpt-3.5-turbo\footnote{https://api.openai.com/v1/chat/completions}) and Claude (claude-1\footnote{https://api.anthropic.com/v1/complete}). Experiments were conducted in June 2023.

\subsection{Benchmarks}
For benchmarks, we used NQ \citep{kwiatkowski-etal-2019-natural}, HotPotQA \citep{yang-etal-2018-hotpotqa}, StrategyQA \citep{geva-etal-2021-aristotle}, and QASC \citep{Khot2019QASCAD}.
NQ consists of factual questions which can be answered with a single passage. Unlike NQ, HotPotQA consists of questions that require reasoning across multiple passages, known as multi-hop QA. StrategyQA and QASC further extend this multi-hop setting by requiring more implicit reasoning.

For gold context \(c_{gold}\), we use the paragraph(s) from Wikipedia, which are part of the annotations in NQ, HotPotQA, and StrategyQA. For QASC, where such paragraph does not exist, we treat the seed facts that were used to create questions as gold context.
In multi-hop QA, we concatenate all the contexts into a single context. See Appendix~\ref{sec:gold} for details.

\subsection{Results}
We report the results in Table \ref{tab:mainexp}. Paraphrased context outperforms the original gold context for most cases in NQ, HotpotQA, and StrategyQA. This means that paraphrased passages were more helpful than the gold passages, implying the existence of knowledge corpus error. Moreover, using the context paraphrased by different model did not cause any performance depredation, indicating some level of universality in the helpfulness of the passages. We provide further analysis of this finding in Appendix \ref{sec:diffmodel}.

\textbf{QASC}. We attribute the degradation in QASC for two reasons. First, the seed facts, which we considered as gold contexts in QASC, are not from a raw corpus. The seed facts are manually selected from cleaned knowledge sources like WorldTree corpus. This is problematic as the gold contexts we are using represent the best-case scenario in retrieval, thereby eliminating any retrieval error. Second, distractor options in multiple-choice question confuses the model to generate a passage relevant to those options. This results in a passage containing distracting information for answering the question. Examples in Table \ref{tab:qasc} illustrate these two points well.

\begin{table*}[t]
\centering
\small
\begin{tabular}{p{2cm}|p{2.5cm}|p{5cm}|p{5cm}}
\hline
Source & Question / Answer & Gold & Paraphrased \\ 
\hline \hline
Increased focus on the question
& Q. who plays charles on have and have nots \newline A. Nick Sager
& Title: The Haves and the Have Nots (TV series) \newline
The Road to Hell", it was revealed that David had a pulse while Erica's fate was a casualty and died from the explosion. Her death was the fuel for David’s warpath of vengeance on Veronica. - \textbf{Charles Frederickson (Nick Sager)}: (Season 4–present) The president-elect and the love interest to Candace, whom she uses to help her get her money back. (...) & \textbf{Charles, played by Nick Sager}, is one of the main characters on the show The Haves and the Have Nots. He is portrayed as the president-elect and love interest of the character Candace. (...) \\
\hline
Chain-of-thought 
& Q. The VMAQT-1 logo is a female spirit in Irish mythology who heralds the death of what? \newline A. a family member
& Title: VMAQT-1 \newline
(...) The \textbf{VMAQT-1 logo is the Banshee}, an Irish mythological figure foretelling death. Its motto is “Tairngreacht Bas,” Gaelic for “Death Foretold.” 
\newline Title: Banshee \newline
A banshee ( ; Modern Irish "bean sí", from Old Irish: "ban síde" , ] , "woman of the fairy mound" or "fairy woman") is a female spirit in Irish mythology who heralds the \textbf{death of a family member}, usually by shrieking or keening. (...) 
& \textbf{The VMAQT-1 logo features a banshee}, which is a female spirit in Irish mythology. According to Irish mythology, banshees shriek or keen to foretell the \textbf{death of a family member}. (...) \\ 
\hline
Incorporation of \newline commonsense knowledge 
& Q. Was Bruce Lee absent from the 1964 University of Washington graduation ceremony? \newline A. True
& Title: Bruce Lee \newline
Lee is best known as a martial artist, but he also studied drama and Asian and Western philosophy while a student at the University of Washington and throughout his life. (...)
\newline Title: Bruce Lee \newline
Lee \textbf{dropped out of college in early 1964} and moved to Oakland to live with James Yimm Lee. (...) 
& There is no mention of him attending or participating in the graduation in either document. The documents instead mention that in early 1964, Lee dropped out of college and moved to Oakland (...) While this evidence is not definitive, it suggests that \textbf{Lee probably did not attend} or participate in the University of Washington's 1964 graduation ceremony. \\
\hline
\end{tabular}
\caption{Qualitative analysis on the source of knowledge corpus error. Examples are respectively from NQ, HotPotQA, and StrategyQA (from top to bottom). All paraphrases are done by Claude. The \textbf{bold} part highlights how paraphrased context may be more helpful than gold context. See \ref{body:quali} for detailed explanation.}
\label{tab:quali}
\end{table*}

\subsection{Qualitative Analysis}
\label{body:quali}
After manually examining a sample of results from the empirical study, we identify three common factors contributing to knowledge corpus error.

\paragraph{1. Increased focus on the question}
Gold passages are a very small subset of facts from Wikipedia, with the communicative intent to generally inform about the subject. Therefore, gold passages inevitably include information that is irrelevant to the question. LLMs can only filter the helpful information from the gold passage during the paraphrasing. In fact, it has been shown that when both retrieved and generated passages contain the correct answer, the FiD reader can produce more correct answers when reading the generated passages \citep{yu2023generate}. And furthermore, models are sensitive to related but irrelevant information in a phenomena called \textit{damaging retrieval} \citep{10.1145/3477495.3532034, oh-thorne-2023-detrimental}. 
Query-focused paraphrasing acts as an information filter mitigating damaging retrieval.

\paragraph{2. Chain-of-thought}
Some questions require a composition of facts to answer the question. In such case, we observe that paraphrasing is acting in a manner similar to chain-of-thought \citep{wei_chain--thought_2022}. This also highlights the inherent limit of corpus such as Wikipedia, where the explicit composition of information is seldom given.

In the second example of Table \ref{tab:quali}, the question requires combining two distinct facts, one about the military unit (VMAQT-1) and another about Irish mythology (Banshee). The paraphrased context acts somewhat akin to chain-of-thought, resulting in a more helpful context.

\paragraph{3. Incorporation of commonsense knowledge}
Commonsense knowledge plays a crucial role in understanding the world \citep{davis-marcus-commonsense}, but not often explicitly stated, especially in a corpus such as Wikipedia. Language models are known to possess a degree of tacit knowledge \citep{petroni-etal-2019-language}, which can be utilized by knowledge generation \citep{liu-etal-2022-generated}. We observe that during paraphrasing, commonsense knowledge is elicited, aiding the reader.

The third example of Table \ref{tab:quali} illustrates how commonsense knowledge — \textit{someone who dropped out of college will probably not attend graduation ceremony} — is induced during paraphrasing.

\section{Conclusion}
In this work, we demonstrate that generated contexts may be more helpful than retrieved contexts in open-domain question answering. By revisiting the formulation of question answering, we identify a gap where retriever inevitably ignores potentially helpful contexts outside of the corpus. We call this \textit{knowledge corpus error}, and design an experiment in order to observe knowledge corpus error empirically. Paraphrasing the human-annotated gold contexts with LLMs led to increased reader performance in 3 out of 4 QA benchmarks, implying the existence of knowledge corpus error.

\section*{Acknowledgments}
This work was supported by Institute of Information \& communications Technology Planning \& Evaluation (IITP) grant funded by the Korea government (MSIT) (No.2019-0-00075, Artificial Intelligence Graduate School Program (KAIST)) and Artificial intelligence industrial convergence cluster development project funded by the Ministry of Science and ICT(MSIT, Korea) \& Gwangju Metropolitan City.

\section*{Limitations}
The first limitation of this work is that it did not employ it in a retrieval setting. We used a gold context, which we assume is the best-case scenario in retrieval. However, retrieved contexts in real retrieval setting \citep{karpukhin-etal-2020-dense}, i.e., contexts in knowledge corpus other than the gold context, may deviate significantly from the gold context. Therefore, it is hard to discuss the effect of paraphrasing and the degree of knowledge corpus within retrieval. 

The second limitation of this work is that it did not address the practical way to marry retrieval and generation via LLMs. Regardless of the seeming benefit of context generation, this approach suffers from issues such as information staleness and hallucination. Contemporaneous works explore various methods to leverage benefits of both retrieval and generation \citep{he2022rethinking, jiang2023active, xu2023recomp}. This work is primarily concerned with the analytic understanding of how generation may have advantages over retrieval. We believe our work can inspire future contributions on empirical method of incorporating retrieval and generation. 

The third limitation of this work is that its scope was limited to question answering. Conditioning generation on retrieved context is a well-studied approach in language modeling \citep{Khandelwal2020Generalization, yogatama-etal-2021-adaptive, borgeaud22a}. It will be worth exploring how knowledge corpus error manifests within language modeling.

\bibliography{anthology, custom}

\begin{thebibliography}{48}
\expandafter\ifx\csname natexlab\endcsname\relax\def\natexlab#1{#1}\fi

\bibitem[{Borgeaud et~al.(2022)Borgeaud, Mensch, Hoffmann, Cai, Rutherford,
  Millican, Van Den~Driessche, Lespiau, Damoc, Clark, De~Las~Casas, Guy,
  Menick, Ring, Hennigan, Huang, Maggiore, Jones, Cassirer, Brock, Paganini,
  Irving, Vinyals, Osindero, Simonyan, Rae, Elsen, and Sifre}]{borgeaud22a}
Sebastian Borgeaud, Arthur Mensch, Jordan Hoffmann, Trevor Cai, Eliza
  Rutherford, Katie Millican, George~Bm Van Den~Driessche, Jean-Baptiste
  Lespiau, Bogdan Damoc, Aidan Clark, Diego De~Las~Casas, Aurelia Guy, Jacob
  Menick, Roman Ring, Tom Hennigan, Saffron Huang, Loren Maggiore, Chris Jones,
  Albin Cassirer, Andy Brock, Michela Paganini, Geoffrey Irving, Oriol Vinyals,
  Simon Osindero, Karen Simonyan, Jack Rae, Erich Elsen, and Laurent Sifre.
  2022.
\newblock \href {https://proceedings.mlr.press/v162/borgeaud22a.html}
  {Improving language models by retrieving from trillions of tokens}.
\newblock In \emph{Proceedings of the 39th International Conference on Machine
  Learning}, volume 162 of \emph{Proceedings of Machine Learning Research},
  pages 2206--2240. PMLR.

\bibitem[{Brown et~al.(2020)Brown, Mann, Ryder, Subbiah, Kaplan, Dhariwal,
  Neelakantan, Shyam, Sastry, Askell, Agarwal, Herbert-Voss, Krueger, Henighan,
  Child, Ramesh, Ziegler, Wu, Winter, Hesse, Chen, Sigler, Litwin, Gray, Chess,
  Clark, Berner, McCandlish, Radford, Sutskever, and
  Amodei}]{brown_language_2020}
Tom Brown, Benjamin Mann, Nick Ryder, Melanie Subbiah, Jared~D Kaplan, Prafulla
  Dhariwal, Arvind Neelakantan, Pranav Shyam, Girish Sastry, Amanda Askell,
  Sandhini Agarwal, Ariel Herbert-Voss, Gretchen Krueger, Tom Henighan, Rewon
  Child, Aditya Ramesh, Daniel Ziegler, Jeffrey Wu, Clemens Winter, Chris
  Hesse, Mark Chen, Eric Sigler, Mateusz Litwin, Scott Gray, Benjamin Chess,
  Jack Clark, Christopher Berner, Sam McCandlish, Alec Radford, Ilya Sutskever,
  and Dario Amodei. 2020.
\newblock \href
  {https://proceedings.neurips.cc/paper/2020/file/1457c0d6bfcb4967418bfb8ac142f64a-Paper.pdf}
  {Language {Models} are {Few}-{Shot} {Learners}}.
\newblock In \emph{Advances in {Neural} {Information} {Processing} {Systems}},
  volume~33, pages 1877--1901. Curran Associates, Inc.

\bibitem[{Chen et~al.(2017)Chen, Fisch, Weston, and
  Bordes}]{chen-etal-2017-reading}
Danqi Chen, Adam Fisch, Jason Weston, and Antoine Bordes. 2017.
\newblock \href {https://doi.org/10.18653/v1/P17-1171} {Reading {W}ikipedia to
  answer open-domain questions}.
\newblock In \emph{Proceedings of the 55th Annual Meeting of the Association
  for Computational Linguistics (Volume 1: Long Papers)}, pages 1870--1879,
  Vancouver, Canada. Association for Computational Linguistics.

\bibitem[{Chiang et~al.(2023)Chiang, Li, Lin, Sheng, Wu, Zhang, Zheng, Zhuang,
  Zhuang, Gonzalez, Stoica, and Xing}]{vicuna2023}
Wei-Lin Chiang, Zhuohan Li, Zi~Lin, Ying Sheng, Zhanghao Wu, Hao Zhang, Lianmin
  Zheng, Siyuan Zhuang, Yonghao Zhuang, Joseph~E. Gonzalez, Ion Stoica, and
  Eric~P. Xing. 2023.
\newblock \href {https://lmsys.org/blog/2023-03-30-vicuna/} {Vicuna: An
  open-source chatbot impressing gpt-4 with 90\%* chatgpt quality}.

\bibitem[{Davis and Marcus(2015)}]{davis-marcus-commonsense}
Ernest Davis and Gary Marcus. 2015.
\newblock \href {https://doi.org/10.1145/2701413} {Commonsense reasoning and
  commonsense knowledge in artificial intelligence}.
\newblock \emph{Commun. ACM}, 58(9):92–103.

\bibitem[{Geng et~al.(2023)Geng, Gudibande, Liu, Wallace, Abbeel, Levine, and
  Song}]{koala_blogpost_2023}
Xinyang Geng, Arnav Gudibande, Hao Liu, Eric Wallace, Pieter Abbeel, Sergey
  Levine, and Dawn Song. 2023.
\newblock \href {https://bair.berkeley.edu/blog/2023/04/03/koala/} {Koala: A
  dialogue model for academic research}.
\newblock Blog post.

\bibitem[{Geva et~al.(2021)Geva, Khashabi, Segal, Khot, Roth, and
  Berant}]{geva-etal-2021-aristotle}
Mor Geva, Daniel Khashabi, Elad Segal, Tushar Khot, Dan Roth, and Jonathan
  Berant. 2021.
\newblock \href {https://doi.org/10.1162/tacl_a_00370} {Did aristotle use a
  laptop? a question answering benchmark with implicit reasoning strategies}.
\newblock \emph{Transactions of the Association for Computational Linguistics},
  9:346--361.

\bibitem[{Gudibande et~al.(2023)Gudibande, Wallace, Snell, Geng, Liu, Abbeel,
  Levine, and Song}]{gudibande2023false}
Arnav Gudibande, Eric Wallace, Charlie Snell, Xinyang Geng, Hao Liu, Pieter
  Abbeel, Sergey Levine, and Dawn Song. 2023.
\newblock \href {http://arxiv.org/abs/2305.15717} {The false promise of
  imitating proprietary llms}.

\bibitem[{Guu et~al.(2020)Guu, Lee, Tung, Pasupat, and
  Chang}]{guu_retrieval_2020}
Kelvin Guu, Kenton Lee, Zora Tung, Panupong Pasupat, and Mingwei Chang. 2020.
\newblock \href {https://proceedings.mlr.press/v119/guu20a.html} {Retrieval
  {Augmented} {Language} {Model} {Pre}-{Training}}.
\newblock In \emph{Proceedings of the 37th {International} {Conference} on
  {Machine} {Learning}}, pages 3929--3938. PMLR.
\newblock ISSN: 2640-3498.

\bibitem[{Haluptzok et~al.(2023)Haluptzok, Bowers, and
  Kalai}]{haluptzok2023language}
Patrick Haluptzok, Matthew Bowers, and Adam~Tauman Kalai. 2023.
\newblock \href {https://openreview.net/forum?id=SaRj2ka1XZ3} {Language models
  can teach themselves to program better}.
\newblock In \emph{The Eleventh International Conference on Learning
  Representations}.

\bibitem[{He et~al.(2022)He, Zhang, and Roth}]{he2022rethinking}
Hangfeng He, Hongming Zhang, and Dan Roth. 2022.
\newblock \href {http://arxiv.org/abs/2301.00303} {Rethinking with retrieval:
  Faithful large language model inference}.

\bibitem[{Ho et~al.(2023)Ho, Schmid, and Yun}]{ho2023large}
Namgyu Ho, Laura Schmid, and Se-Young Yun. 2023.
\newblock \href {http://arxiv.org/abs/2212.10071} {Large language models are
  reasoning teachers}.

\bibitem[{Honovich et~al.(2022)Honovich, Scialom, Levy, and
  Schick}]{honovich2022unnatural}
Or~Honovich, Thomas Scialom, Omer Levy, and Timo Schick. 2022.
\newblock \href {https://arxiv.org/abs/2212.09689} {Unnatural instructions:
  Tuning language models with (almost) no human labor}.

\bibitem[{Izacard and Grave(2021)}]{izacard-grave-2021-leveraging}
Gautier Izacard and Edouard Grave. 2021.
\newblock \href {https://doi.org/10.18653/v1/2021.eacl-main.74} {Leveraging
  passage retrieval with generative models for open domain question answering}.
\newblock In \emph{Proceedings of the 16th Conference of the European Chapter
  of the Association for Computational Linguistics: Main Volume}, pages
  874--880, Online. Association for Computational Linguistics.

\bibitem[{Jiang et~al.(2023)Jiang, Xu, Gao, Sun, Liu, Dwivedi-Yu, Yang, Callan,
  and Neubig}]{jiang2023active}
Zhengbao Jiang, Frank~F. Xu, Luyu Gao, Zhiqing Sun, Qian Liu, Jane Dwivedi-Yu,
  Yiming Yang, Jamie Callan, and Graham Neubig. 2023.
\newblock \href {http://arxiv.org/abs/2305.06983} {Active retrieval augmented
  generation}.

\bibitem[{Kaplan et~al.(2020)Kaplan, McCandlish, Henighan, Brown, Chess, Child,
  Gray, Radford, Wu, and Amodei}]{kaplan2020scaling}
Jared Kaplan, Sam McCandlish, Tom Henighan, Tom~B. Brown, Benjamin Chess, Rewon
  Child, Scott Gray, Alec Radford, Jeffrey Wu, and Dario Amodei. 2020.
\newblock \href {http://arxiv.org/abs/2001.08361} {Scaling laws for neural
  language models}.

\bibitem[{Karpukhin et~al.(2020)Karpukhin, Oguz, Min, Lewis, Wu, Edunov, Chen,
  and Yih}]{karpukhin-etal-2020-dense}
Vladimir Karpukhin, Barlas Oguz, Sewon Min, Patrick Lewis, Ledell Wu, Sergey
  Edunov, Danqi Chen, and Wen-tau Yih. 2020.
\newblock \href {https://doi.org/10.18653/v1/2020.emnlp-main.550} {Dense
  passage retrieval for open-domain question answering}.
\newblock In \emph{Proceedings of the 2020 Conference on Empirical Methods in
  Natural Language Processing (EMNLP)}, pages 6769--6781, Online. Association
  for Computational Linguistics.

\bibitem[{Khandelwal et~al.(2020)Khandelwal, Levy, Jurafsky, Zettlemoyer, and
  Lewis}]{Khandelwal2020Generalization}
Urvashi Khandelwal, Omer Levy, Dan Jurafsky, Luke Zettlemoyer, and Mike Lewis.
  2020.
\newblock \href {https://openreview.net/forum?id=HklBjCEKvH} {Generalization
  through memorization: Nearest neighbor language models}.
\newblock In \emph{International Conference on Learning Representations}.

\bibitem[{Khot et~al.(2019)Khot, Clark, Guerquin, Jansen, and
  Sabharwal}]{Khot2019QASCAD}
Tushar Khot, Peter Clark, Michal Guerquin, Peter~Alexander Jansen, and Ashish
  Sabharwal. 2019.
\newblock Qasc: A dataset for question answering via sentence composition.
\newblock In \emph{AAAI Conference on Artificial Intelligence}.

\bibitem[{Kojima et~al.(2022)Kojima, Gu, Reid, Matsuo, and
  Iwasawa}]{kojima2022}
Takeshi Kojima, Shixiang~(Shane) Gu, Machel Reid, Yutaka Matsuo, and Yusuke
  Iwasawa. 2022.
\newblock \href
  {https://proceedings.neurips.cc/paper_files/paper/2022/file/8bb0d291acd4acf06ef112099c16f326-Paper-Conference.pdf}
  {Large language models are zero-shot reasoners}.
\newblock In \emph{Advances in Neural Information Processing Systems},
  volume~35, pages 22199--22213. Curran Associates, Inc.

\bibitem[{Kwiatkowski et~al.(2019)Kwiatkowski, Palomaki, Redfield, Collins,
  Parikh, Alberti, Epstein, Polosukhin, Devlin, Lee, Toutanova, Jones, Kelcey,
  Chang, Dai, Uszkoreit, Le, and Petrov}]{kwiatkowski-etal-2019-natural}
Tom Kwiatkowski, Jennimaria Palomaki, Olivia Redfield, Michael Collins, Ankur
  Parikh, Chris Alberti, Danielle Epstein, Illia Polosukhin, Jacob Devlin,
  Kenton Lee, Kristina Toutanova, Llion Jones, Matthew Kelcey, Ming-Wei Chang,
  Andrew~M. Dai, Jakob Uszkoreit, Quoc Le, and Slav Petrov. 2019.
\newblock \href {https://doi.org/10.1162/tacl_a_00276} {Natural questions: A
  benchmark for question answering research}.
\newblock \emph{Transactions of the Association for Computational Linguistics},
  7:452--466.

\bibitem[{Lee et~al.(2019)Lee, Chang, and Toutanova}]{lee-etal-2019-latent}
Kenton Lee, Ming-Wei Chang, and Kristina Toutanova. 2019.
\newblock \href {https://doi.org/10.18653/v1/P19-1612} {Latent retrieval for
  weakly supervised open domain question answering}.
\newblock In \emph{Proceedings of the 57th Annual Meeting of the Association
  for Computational Linguistics}, pages 6086--6096, Florence, Italy.
  Association for Computational Linguistics.

\bibitem[{Lewis et~al.(2020)Lewis, Perez, Piktus, Petroni, Karpukhin, Goyal,
  K\"{u}ttler, Lewis, Yih, Rockt\"{a}schel, Riedel, and Kiela}]{lewis2020rag}
Patrick Lewis, Ethan Perez, Aleksandra Piktus, Fabio Petroni, Vladimir
  Karpukhin, Naman Goyal, Heinrich K\"{u}ttler, Mike Lewis, Wen-tau Yih, Tim
  Rockt\"{a}schel, Sebastian Riedel, and Douwe Kiela. 2020.
\newblock \href
  {https://proceedings.neurips.cc/paper_files/paper/2020/file/6b493230205f780e1bc26945df7481e5-Paper.pdf}
  {Retrieval-augmented generation for knowledge-intensive nlp tasks}.
\newblock In \emph{Advances in Neural Information Processing Systems},
  volume~33, pages 9459--9474. Curran Associates, Inc.

\bibitem[{Liu et~al.(2022{\natexlab{a}})Liu, Swayamdipta, Smith, and
  Choi}]{liu-etal-2022-wanli}
Alisa Liu, Swabha Swayamdipta, Noah~A. Smith, and Yejin Choi.
  2022{\natexlab{a}}.
\newblock \href {https://aclanthology.org/2022.findings-emnlp.508} {{WANLI}:
  Worker and {AI} collaboration for natural language inference dataset
  creation}.
\newblock In \emph{Findings of the Association for Computational Linguistics:
  EMNLP 2022}, pages 6826--6847, Abu Dhabi, United Arab Emirates. Association
  for Computational Linguistics.

\bibitem[{Liu et~al.(2022{\natexlab{b}})Liu, Liu, Lu, Welleck, West, Le~Bras,
  Choi, and Hajishirzi}]{liu-etal-2022-generated}
Jiacheng Liu, Alisa Liu, Ximing Lu, Sean Welleck, Peter West, Ronan Le~Bras,
  Yejin Choi, and Hannaneh Hajishirzi. 2022{\natexlab{b}}.
\newblock \href {https://doi.org/10.18653/v1/2022.acl-long.225} {Generated
  knowledge prompting for commonsense reasoning}.
\newblock In \emph{Proceedings of the 60th Annual Meeting of the Association
  for Computational Linguistics (Volume 1: Long Papers)}, pages 3154--3169,
  Dublin, Ireland. Association for Computational Linguistics.

\bibitem[{Magister et~al.(2023)Magister, Mallinson, Adamek, Malmi, and
  Severyn}]{magister2023teaching}
Lucie~Charlotte Magister, Jonathan Mallinson, Jakub Adamek, Eric Malmi, and
  Aliaksei Severyn. 2023.
\newblock \href {http://arxiv.org/abs/2212.08410} {Teaching small language
  models to reason}.

\bibitem[{Oh and Thorne(2023)}]{oh-thorne-2023-detrimental}
Philhoon Oh and James Thorne. 2023.
\newblock Detrimental contexts in open-domain question answering.
\newblock In \emph{Findings of the Association for Computational Linguistics:
  EMNLP 2023}, Singapore. Association for Computational Linguistics.

\bibitem[{Petroni et~al.(2021)Petroni, Piktus, Fan, Lewis, Yazdani, De~Cao,
  Thorne, Jernite, Karpukhin, Maillard, Plachouras, Rockt{\"a}schel, and
  Riedel}]{petroni-etal-2021-kilt}
Fabio Petroni, Aleksandra Piktus, Angela Fan, Patrick Lewis, Majid Yazdani,
  Nicola De~Cao, James Thorne, Yacine Jernite, Vladimir Karpukhin, Jean
  Maillard, Vassilis Plachouras, Tim Rockt{\"a}schel, and Sebastian Riedel.
  2021.
\newblock \href {https://doi.org/10.18653/v1/2021.naacl-main.200} {{KILT}: a
  benchmark for knowledge intensive language tasks}.
\newblock In \emph{Proceedings of the 2021 Conference of the North American
  Chapter of the Association for Computational Linguistics: Human Language
  Technologies}, pages 2523--2544, Online. Association for Computational
  Linguistics.

\bibitem[{Petroni et~al.(2019)Petroni, Rockt{\"a}schel, Riedel, Lewis, Bakhtin,
  Wu, and Miller}]{petroni-etal-2019-language}
Fabio Petroni, Tim Rockt{\"a}schel, Sebastian Riedel, Patrick Lewis, Anton
  Bakhtin, Yuxiang Wu, and Alexander Miller. 2019.
\newblock \href {https://doi.org/10.18653/v1/D19-1250} {Language models as
  knowledge bases?}
\newblock In \emph{Proceedings of the 2019 Conference on Empirical Methods in
  Natural Language Processing and the 9th International Joint Conference on
  Natural Language Processing (EMNLP-IJCNLP)}, pages 2463--2473, Hong Kong,
  China. Association for Computational Linguistics.

\bibitem[{Raffel et~al.(2020)Raffel, Shazeer, Roberts, Lee, Narang, Matena,
  Zhou, Li, and Liu}]{2020t5}
Colin Raffel, Noam Shazeer, Adam Roberts, Katherine Lee, Sharan Narang, Michael
  Matena, Yanqi Zhou, Wei Li, and Peter~J. Liu. 2020.
\newblock \href {http://jmlr.org/papers/v21/20-074.html} {Exploring the limits
  of transfer learning with a unified text-to-text transformer}.
\newblock \emph{Journal of Machine Learning Research}, 21(140):1--67.

\bibitem[{Roberts et~al.(2020)Roberts, Raffel, and
  Shazeer}]{roberts-etal-2020-much}
Adam Roberts, Colin Raffel, and Noam Shazeer. 2020.
\newblock \href {https://doi.org/10.18653/v1/2020.emnlp-main.437} {How much
  knowledge can you pack into the parameters of a language model?}
\newblock In \emph{Proceedings of the 2020 Conference on Empirical Methods in
  Natural Language Processing (EMNLP)}, pages 5418--5426, Online. Association
  for Computational Linguistics.

\bibitem[{Sanh et~al.(2022)Sanh, Webson, Raffel, Bach, Sutawika, Alyafeai,
  Chaffin, Stiegler, Raja, Dey, Bari, Xu, Thakker, Sharma, Szczechla, Kim,
  Chhablani, Nayak, Datta, Chang, Jiang, Wang, Manica, Shen, Yong, Pandey,
  Bawden, Wang, Neeraj, Rozen, Sharma, Santilli, Fevry, Fries, Teehan, Scao,
  Biderman, Gao, Wolf, and Rush}]{sanh2022multitask}
Victor Sanh, Albert Webson, Colin Raffel, Stephen Bach, Lintang Sutawika, Zaid
  Alyafeai, Antoine Chaffin, Arnaud Stiegler, Arun Raja, Manan Dey, M~Saiful
  Bari, Canwen Xu, Urmish Thakker, Shanya~Sharma Sharma, Eliza Szczechla,
  Taewoon Kim, Gunjan Chhablani, Nihal Nayak, Debajyoti Datta, Jonathan Chang,
  Mike Tian-Jian Jiang, Han Wang, Matteo Manica, Sheng Shen, Zheng~Xin Yong,
  Harshit Pandey, Rachel Bawden, Thomas Wang, Trishala Neeraj, Jos Rozen,
  Abheesht Sharma, Andrea Santilli, Thibault Fevry, Jason~Alan Fries, Ryan
  Teehan, Teven~Le Scao, Stella Biderman, Leo Gao, Thomas Wolf, and Alexander~M
  Rush. 2022.
\newblock \href {https://openreview.net/forum?id=9Vrb9D0WI4} {Multitask
  prompted training enables zero-shot task generalization}.
\newblock In \emph{International Conference on Learning Representations}.

\bibitem[{Sauchuk et~al.(2022)Sauchuk, Thorne, Halevy, Tonellotto, and
  Silvestri}]{10.1145/3477495.3532034}
Artsiom Sauchuk, James Thorne, Alon Halevy, Nicola Tonellotto, and Fabrizio
  Silvestri. 2022.
\newblock \href {https://doi.org/10.1145/3477495.3532034} {On the role of
  relevance in natural language processing tasks}.
\newblock In \emph{Proceedings of the 45th International ACM SIGIR Conference
  on Research and Development in Information Retrieval}, SIGIR '22, page
  1785–1789, New York, NY, USA. Association for Computing Machinery.

\bibitem[{Singh et~al.(2021)Singh, Reddy, Hamilton, Dyer, and
  Yogatama}]{singh_end--end_2021}
Devendra Singh, Siva Reddy, Will Hamilton, Chris Dyer, and Dani Yogatama. 2021.
\newblock \href
  {https://papers.nips.cc/paper_files/paper/2021/hash/da3fde159d754a2555eaa198d2d105b2-Abstract.html}
  {End-to-{End} {Training} of {Multi}-{Document} {Reader} and {Retriever} for
  {Open}-{Domain} {Question} {Answering}}.
\newblock In \emph{Advances in {Neural} {Information} {Processing} {Systems}},
  volume~34, pages 25968--25981. Curran Associates, Inc.

\bibitem[{Sun et~al.(2023)Sun, Wang, Tay, Yang, and
  Zhou}]{sun2023recitationaugmented}
Zhiqing Sun, Xuezhi Wang, Yi~Tay, Yiming Yang, and Denny Zhou. 2023.
\newblock \href {https://openreview.net/forum?id=-cqvvvb-NkI}
  {Recitation-augmented language models}.
\newblock In \emph{The Eleventh International Conference on Learning
  Representations}.

\bibitem[{Taori et~al.(2023)Taori, Gulrajani, Zhang, Dubois, Li, Guestrin,
  Liang, and Hashimoto}]{alpaca}
Rohan Taori, Ishaan Gulrajani, Tianyi Zhang, Yann Dubois, Xuechen Li, Carlos
  Guestrin, Percy Liang, and Tatsunori~B. Hashimoto. 2023.
\newblock Stanford alpaca: An instruction-following llama model.
\newblock \url{https://github.com/tatsu-lab/stanford_alpaca}.

\bibitem[{Thorne et~al.(2018)Thorne, Vlachos, Christodoulopoulos, and
  Mittal}]{thorne-etal-2018-fever}
James Thorne, Andreas Vlachos, Christos Christodoulopoulos, and Arpit Mittal.
  2018.
\newblock \href {https://doi.org/10.18653/v1/N18-1074} {{FEVER}: a large-scale
  dataset for fact extraction and {VER}ification}.
\newblock In \emph{Proceedings of the 2018 Conference of the North {A}merican
  Chapter of the Association for Computational Linguistics: Human Language
  Technologies, Volume 1 (Long Papers)}, pages 809--819, New Orleans,
  Louisiana. Association for Computational Linguistics.

\bibitem[{Touvron et~al.(2023)Touvron, Lavril, Izacard, Martinet, Lachaux,
  Lacroix, Rozi{\`e}re, Goyal, Hambro, Azhar, Rodriguez, Joulin, Grave, and
  Lample}]{touvron2023llama}
Hugo Touvron, Thibaut Lavril, Gautier Izacard, Xavier Martinet, Marie-Anne
  Lachaux, Timoth{\'e}e Lacroix, Baptiste Rozi{\`e}re, Naman Goyal, Eric
  Hambro, Faisal Azhar, Aurelien Rodriguez, Armand Joulin, Edouard Grave, and
  Guillaume Lample. 2023.
\newblock Llama: Open and efficient foundation language models.
\newblock \emph{arXiv preprint arXiv:2302.13971}.

\bibitem[{Wang et~al.(2023)Wang, Wei, Schuurmans, Le, Chi, Narang, Chowdhery,
  and Zhou}]{wang2023selfconsistency}
Xuezhi Wang, Jason Wei, Dale Schuurmans, Quoc~V Le, Ed~H. Chi, Sharan Narang,
  Aakanksha Chowdhery, and Denny Zhou. 2023.
\newblock \href {https://openreview.net/forum?id=1PL1NIMMrw} {Self-consistency
  improves chain of thought reasoning in language models}.
\newblock In \emph{The Eleventh International Conference on Learning
  Representations}.

\bibitem[{Wang et~al.(2022)Wang, Kordi, Mishra, Liu, Smith, Khashabi, and
  Hajishirzi}]{wang_self-instruct_2023}
Yizhong Wang, Yeganeh Kordi, Swaroop Mishra, Alisa Liu, Noah~A. Smith, Daniel
  Khashabi, and Hannaneh Hajishirzi. 2022.
\newblock \href {https://arxiv.org/abs/2212.10560} {Self-instruct: Aligning
  language model with self generated instructions}.

\bibitem[{Wei et~al.(2022{\natexlab{a}})Wei, Bosma, Zhao, Guu, Yu, Lester, Du,
  Dai, and Le}]{wei2022finetuned}
Jason Wei, Maarten Bosma, Vincent Zhao, Kelvin Guu, Adams~Wei Yu, Brian Lester,
  Nan Du, Andrew~M. Dai, and Quoc~V Le. 2022{\natexlab{a}}.
\newblock \href {https://openreview.net/forum?id=gEZrGCozdqR} {Finetuned
  language models are zero-shot learners}.
\newblock In \emph{International Conference on Learning Representations}.

\bibitem[{Wei et~al.(2022{\natexlab{b}})Wei, Wang, Schuurmans, Bosma, Ichter,
  Xia, Chi, Le, and Zhou}]{wei_chain--thought_2022}
Jason Wei, Xuezhi Wang, Dale Schuurmans, Maarten Bosma, Brian Ichter, Fei Xia,
  Ed~Chi, Quoc~V. Le, and Denny Zhou. 2022{\natexlab{b}}.
\newblock \href
  {https://papers.nips.cc/paper_files/paper/2022/hash/9d5609613524ecf4f15af0f7b31abca4-Abstract-Conference.html}
  {Chain-of-{Thought} {Prompting} {Elicits} {Reasoning} in {Large} {Language}
  {Models}}.
\newblock \emph{Advances in Neural Information Processing Systems},
  35:24824--24837.

\bibitem[{Xu et~al.(2023)Xu, Shi, and Choi}]{xu2023recomp}
Fangyuan Xu, Weijia Shi, and Eunsol Choi. 2023.
\newblock \href {http://arxiv.org/abs/2310.04408} {Recomp: Improving
  retrieval-augmented lms with compression and selective augmentation}.

\bibitem[{Yang et~al.(2018)Yang, Qi, Zhang, Bengio, Cohen, Salakhutdinov, and
  Manning}]{yang-etal-2018-hotpotqa}
Zhilin Yang, Peng Qi, Saizheng Zhang, Yoshua Bengio, William Cohen, Ruslan
  Salakhutdinov, and Christopher~D. Manning. 2018.
\newblock \href {https://doi.org/10.18653/v1/D18-1259} {{H}otpot{QA}: A dataset
  for diverse, explainable multi-hop question answering}.
\newblock In \emph{Proceedings of the 2018 Conference on Empirical Methods in
  Natural Language Processing}, pages 2369--2380, Brussels, Belgium.
  Association for Computational Linguistics.

\bibitem[{Ye et~al.(2022)Ye, Gao, Li, Xu, Feng, Wu, Yu, and
  Kong}]{ye-etal-2022-zerogen}
Jiacheng Ye, Jiahui Gao, Qintong Li, Hang Xu, Jiangtao Feng, Zhiyong Wu, Tao
  Yu, and Lingpeng Kong. 2022.
\newblock \href {https://aclanthology.org/2022.emnlp-main.801} {{Z}ero{G}en:
  Efficient zero-shot learning via dataset generation}.
\newblock In \emph{Proceedings of the 2022 Conference on Empirical Methods in
  Natural Language Processing}, pages 11653--11669, Abu Dhabi, United Arab
  Emirates. Association for Computational Linguistics.

\bibitem[{Yogatama et~al.(2021)Yogatama, de~Masson~d{'}Autume, and
  Kong}]{yogatama-etal-2021-adaptive}
Dani Yogatama, Cyprien de~Masson~d{'}Autume, and Lingpeng Kong. 2021.
\newblock \href {https://doi.org/10.1162/tacl_a_00371} {Adaptive semiparametric
  language models}.
\newblock \emph{Transactions of the Association for Computational Linguistics},
  9:362--373.

\bibitem[{Yu et~al.(2023)Yu, Iter, Wang, Xu, Ju, Sanyal, Zhu, Zeng, and
  Jiang}]{yu2023generate}
Wenhao Yu, Dan Iter, Shuohang Wang, Yichong Xu, Mingxuan Ju, Soumya Sanyal,
  Chenguang Zhu, Michael Zeng, and Meng Jiang. 2023.
\newblock \href {https://openreview.net/forum?id=fB0hRu9GZUS} {Generate rather
  than retrieve: Large language models are strong context generators}.
\newblock In \emph{The Eleventh International Conference on Learning
  Representations}.

\bibitem[{Zelikman et~al.(2022)Zelikman, Wu, Mu, and
  Goodman}]{zelikman_star_2022}
Eric Zelikman, Yuhuai Wu, Jesse Mu, and Noah Goodman. 2022.
\newblock \href
  {https://proceedings.neurips.cc/paper_files/paper/2022/file/639a9a172c044fbb64175b5fad42e9a5-Paper-Conference.pdf}
  {{STaR}: {Bootstrapping} {Reasoning} {With} {Reasoning}}.
\newblock In \emph{Advances in {Neural} {Information} {Processing} {Systems}},
  volume~35, pages 15476--15488. Curran Associates, Inc.

\end{thebibliography}
\bibliographystyle{acl_natbib}

\appendix

\section{Dataset}
\label{sec:dataset}

\textbf{NQ}: We use the KILT \citep{petroni-etal-2021-kilt} version of NQ\footnote{https://github.com/facebookresearch/KILT}. We use the dev split after excluding the instances where context does not include answers, which results in 2532 samples. \\
\textbf{HotPotQA}: We use the dataset from its original source\footnote{https://hotpotqa.github.io/}. We use the dev split, which includes 7405 samples. To reduce the inference cost, we only use the subset of first 1531 samples.\\
\textbf{StrategyQA}: We use the dataset from its original source \footnote{https://allenai.org/data/strategyqa}. We use the training split, which includes 2290 samples. We use the training split because the dev split contained too few (490) examples. \\
\textbf{QASC}: We use the dataset from its original source \footnote{https://allenai.org/data/qasc}. We use the dev split, which includes 926 samples.

\section{Detailed experimental setup}
\label{sec:exp}

\subsection{Selection of gold passage}
\label{sec:gold}
\textbf{NQ}: NQ contains a set of provenances for possible answer contexts. For the experiments, we select the gold passages from the provenances that include at least one of the candidate answers. When there are multiple good passages, we employ the very first one. \\
\textbf{HotPotQA}: HotPotQA contains 2 gold paragraphs from Wikipedia for each question. A gold passage is simply the concatenation of these two. Note that we do not utilize fine-grained sentence-level annotation in 2 paragraphs.\\
\textbf{StrategyQA}: StrategyQA contains decomposition steps to solve the question. Each of these steps may be attached with a supporting paragraph from Wikipedia. A gold passage is the concatenation of all these paragraphs throughout the whole steps. Among three different annotated decomposing steps in the dataset, we use the first one.\\ 
\textbf{QASC}: QASC contains two facts that are combined to create a question. These facts are selected from a cleaned knowledge source. A gold passage is simply the concatenation of these two facts.

The title of the passage is prepended to the passage in cases where titles are available (NQ, HotPotQA, and StrategyQA).

\subsection{Details on generation}
\label{sec:gen}
\textbf{NQ}: During reading, we used 3-shot prompting, where the 3-shot demonstrations are sampled from GPT-3.5 with questions from the dev split of NQ. Note that these questions are excluded from the experiment. Max tokens to generate was set to 500 in paraphrase and 25 in read.\\
\textbf{HotPotQA}: Max tokens to generate was set to 300 in paraphrase and 10 in read. \\
\textbf{StrategyQA}: Max tokens to generate was set to 300 in paraphrase and 10 in read. \\
\textbf{QASC}: Max tokens to generate was set to 100 in paraphrase and 10 in read. Temperature during generation was set to 0.8 in paraphrase and 0.4 in read.

We used 3-shot prompting for reading in NQ but otherwise used zero-shot prompting. Other generation keyword arguments are set to default if not specified. For the prompts used, see Table \ref{tab:prompt}.

Hyperparameters related to generation are decided mainly through trial-and-error. For example, max tokens was adjusted according to few preliminary samples. We tried to tweak temperature for QASC after observing deviant result, but only had minor impact. 3-shot setup was chosen for NQ because the performance was too low in zero-shot.

\section{Details on evaluation}
\label{sec:maineval}
Evaluating exact match or accuracy may be non-trivial in a generative setting. Hence, we follow the previous works \citep{kojima2022, yu2023generate}. \\
\textbf{NQ (Exact Match)}: Following \citealt{yu2023generate}, we measure exact match of the output string after normalization. \\
\textbf{HotPotQA (Exact Match)}: Similarly as \citealt{yu2023generate}, we measure exact match of the output string after normalization. \\
\textbf{StrategyQA (Accuracy)}: Following \citealt{kojima2022}, we measure accuracy by picking up the first "yes" or "no" encountered in the text after removing unnecessary letters. \\
\textbf{QASC (Accuracy)}: Similarly as \citealt{kojima2022}, we measure accuracy by picking up the ﬁrst large letter out of A to H encountered in the text.

\section{Inference Cost}
\label{sec:cost}
We used OpenAI and Anthropic's API to use their LLMs. The cost for OpenAI's API is estimated to be around \$40 to \$50. Anthropic's API has not cost any as we were on the free version.

\section{Accordance between heterogeneous reader models}
\label{sec:diffmodel}

\begin{table}[ht]
\small
\centering
\begin{tabular}{c||c|c|c|c}
    \multicolumn{1}{c||}{\multirow{2}{*}{Given context}} & \multicolumn{2}{c|}{Both} & \multicolumn{2}{c}{Only correct for} \\
    \cline{2-5}
    \multicolumn{1}{c||}{} & correct & wrong & GPT & Claude \\
    \hline
    Gold & \multicolumn{1}{c|}{27} & \multicolumn{1}{c|}{43} & \multicolumn{1}{c|}{9} & \multicolumn{1}{c}{21} \\
    \hline
    GPT-paraph & \multicolumn{1}{c|}{33} & \multicolumn{1}{c|}{44} & \multicolumn{1}{c|}{5} & \multicolumn{1}{c}{18} \\
    \hline
    Claude-paraph & \multicolumn{1}{c|}{39} & \multicolumn{1}{c|}{42} & \multicolumn{1}{c|}{4} & \multicolumn{1}{c}{15} \\
\end{tabular}
\caption{Accordance between two heterogeneous reader models, GPT and Claude. The table is based on the reader performance (exact match (\%)) in HotPotQA when given gold context, GPT-paraphrased context, and Claude-paraphrased context (from top to bottom).}
\label{tab:hetero}
\end{table}

For a majority of the examples, two readers accord with each other, i.e., both are correct or wrong, and this ratio is even higher in paraphrased contexts.

\section{Examples}
\label{sec:example}
For examples of paraphrased context, see Table \ref{tab:nq} through \ref{tab:qasc}.

\begin{table*}[t]
\centering
\begin{tabular}{c|p{6cm}p{6cm}}
\hline
Benchmark & Paraphrase & Read \\
\hline
NQ & Paraphrase a background document in your own words to answer the given question. \newline \newline Question: \{q\} \newline \newline Document: \{c\} & Referring to the passage, find the correct answer (just one entity) to the given question. I will first show you few examples. \newline \newline Passage: \{c1\} \newline Question: \{q1\} \newline Answer: \{a1\} \newline  \newline \textit{example continues} \newline \newline Passage: \{c\} \newline Question: \{q\} \newline Answer: \\
\hline
HotPotQA & Paraphrase the background documents into a single document in your own words to answer the given question. \newline \newline Question: \{q\} \newline \newline Documents: \{c\} & Referring to the passage, find the correct answer (just one entity) to the given question. \newline \newline \{c\} \newline \newline \{q\} \\
\hline
StrategyQA & Paraphrase the background documents into a single document in your own words to answer the given question. \newline \newline Question: \{q\} \newline \newline Documents: \{c\} & Read the passage and answer the question with yes or no. \newline \newline \{c\} \newline \newline \{q\}\\
\hline
QASC & Paraphrase the background documents into a single document in your own words to answer the given question. \newline \newline Question: \{q\} \newline \newline Documents: \{c\} & Read the passage and answer the question with one of A, B, C, D, E, F, G, or H. \newline \newline \{c\} \newline \newline \{q\} \\
\hline
\end{tabular}
\caption{\textbf{Prompts used for paraphrasing and reading.} \{q\}, \{c\}, and \{a\} each indicate question, context, and answer from the test instance. \{q1\}, \{c1\}, and \{a1\} indicate one of the demonstrations used for few-shot prompting. Note that the full few-shot prompt format is not displayed for conciseness.}
\label{tab:prompt}
\end{table*}

\begin{table*}
\begin{tabular}{c|p{7cm}p{7cm}}
\hline
Benchmark & Gold & Paraphrased \\
\hline
NQ & 
\textbf{Question: what does the pearl symbolize in the bible} \newline \newline
Title: Parable of the Pearl \newline \newline Parable of the Pearl The Parable of the Pearl (also called the Pearl of Great Price) is one of the parables of Jesus. It appears in and illustrates the great value of \hl{the Kingdom of Heaven}. It is the penultimate parable in Matthew 13, just before the Parable of the Dragnet. It immediately follows the Parable of the Hidden Treasure, which has a similar theme. It does not appear in the other synoptic gospels but a version of this parable does appear in the non-canonical Gospel of Thomas & 
\textbf{GPT: } The Pearl of Great Price is a parable told by Jesus that is found in the book of Matthew. It teaches us about the immense worth of \hl{the Kingdom of Heaven}. This parable is situated just before the Parable of the Dragnet and comes after the Parable of the Hidden Treasure, which has a similar meaning. (...) \newline
\textit{GPT: X -> X, Claude: X -> O} \newline \newline
\textbf{Claude: } In summary, the pearl symbolizes \hl{the Kingdom of Heaven} in the Bible. This is illustrated in the Parable of the Pearl told by Jesus, where a merchant finds a pearl of great value and sells all his possessions so that he can buy it. The pearl represents \hl{the Kingdom of Heaven}, which is so valuable that one should sacrifice everything else to obtain it. \newline
\textit{GPT: X -> O, Claude: X -> X}
\\
\cline{2-3}
& \textbf{Question: when did the song here comes the boom come out} \newline \newline
 Title: Boom (P.O.D. song)\newline\newline Boom (P.O.D. song) "Boom" is a song by American rock band P.O.D. It was released in \hl{May 2002} as the third single from their second major label studio album "Satellite". While it did not chart as well as the album\'s previous singles, the song has appeared significantly in film and television. "Boom (The Crystal Method remix)" was included on the remix album "Community Service" and as a bonus track on the special edition re-release of "Satellite" available August 27 & 
\textbf{GPT: } "Here Comes the Boom" is a song by P.O.D., an American rock band. It was released in \hl{May 2002} as the third single from their album "Satellite." Although it didn't perform as well as the previous singles from the album, it has been featured in numerous movies and TV shows. (...) \newline
\textit{GPT: X -> O, Claude: X -> O} \newline \newline
\textbf{Claude: } The song 'Here Comes the Boom' by the rock band P.O.D. was released in \hl{May 2002}. It was the third single from their album 'Satellite' released that year. Though the song did not chart highly, it has been featured in various films and TV shows. A remixed version of the song by The Crystal Method was included in P.O.D.'s remix album 'Community Service' as well as in a special re-release of the 'Satellite' album in August 2002. \newline
\textit{GPT: X -> O, Claude: X -> X}
\\
\hline
\end{tabular}
\caption{\textbf{Examples of gold contexts and paraphrased contexts in NQ}. Answers are highlighted. Below the paraphrased context are the results of each reader. \textit{GPT: X -> O} indicates that GPT reader got wrong with original context, but right with paraphrased context.}
\label{tab:nq}
\end{table*}

\begin{table*}
\begin{tabular}{c|p{7cm}p{7cm}}
\hline
Benchmark & Gold & Paraphrased \\
\hline
HotPotQA & 
\textbf{Question: The VMAQT-1 logo is a female spirit in Irish mythology who heralds the death of what?} \newline \newline
Title: VMAQT-1\newline \newline Marine Tactical Electronic Warfare Training Squadron 1 (VMAQT-1) is a United States Marine Corps electronic warfare training squadron consisting of EA-6B Prowler jets. The squadron is based at Marine Corps Air Station Cherry Point, North Carolina and falls under the command of Marine Aircraft Group 14 (MAG-14) and the 2nd Marine Aircraft Wing (2nd MAW). The VMAQT-1 logo is the Banshee, an Irish mythological figure foretelling death. Its motto is “Tairngreacht Bas,” Gaelic for “Death Foretold.”\newline \newline Title: Banshee\newline \newline A banshee ( ; Modern Irish "bean sí", from Old Irish: "ban síde" , ] , "woman of the fairy mound" or "fairy woman") is a female spirit in Irish mythology who heralds the death of \hl{a family member}, usually by shrieking or keening. Her name is connected to the mythologically-important tumuli or "mounds" that dot the Irish countryside, which are known as síde (singular "síd") in Old Irish. & 
\textbf{GPT: } VMAQT-1 is a US Marine Corps training squadron consisting of EA-6B Prowler jets based in North Carolina. Their logo is the Banshee, a female spirit from Irish mythology who predicts death in a family by screaming or keening. The Banshee\'s name is associated with the mounds in the Irish countryside known as "síde." The squadron\'s motto is "Tairngreacht Bas," meaning "Death Foretold" in Gaelic. \newline
\textit{GPT: O -> X, Claude: X -> X} \newline \newline
\textbf{Claude: } The VMAQT-1 logo features a banshee, which is a female spirit in Irish mythology. According to Irish mythology, banshees shriek or keen to foretell the death of \hl{a family member}. The banshee's name comes from síde, the mythological mounds in the Irish countryside. Therefore, the banshee in the VMAQT-1 logo heralds the death of a family member. \newline
\textit{GPT: O -> O, Claude: X -> O}
\\
\hline
\end{tabular}
\caption{\textbf{An example of a gold context and paraphrased contexts in HotPotQA.} Answers are highlighted. Below the paraphrased context are the results of each reader. \textit{GPT: X -> O} indicates that GPT reader got wrong with original context, but right with paraphrased context.}
\label{tab:hotpot}
\end{table*}

\begin{table*}
\begin{tabular}{c|p{7cm}p{7cm}}
\hline
Benchmark & Gold & Paraphrased \\
\hline
StrategyQA & 
\textbf{Question: Can Lamborghini's fastest model win a race against a Porsche 911? \newline Answer: yes} \newline \newline
Title: Lamborghini Veneno \newline \newline The Lamborghini Veneno (Spanish pronunciation:[be neno]) is a limited production high performance sportscar manufactured by Italian automobile manufacturer Lamborghini. (...)
The Lamborghini Veneno has a top speed of 356 km/h (221 mph) and has a 0–97 km/h (0–60 mph) acceleration time of 2.8 seconds. The car has a braking distance of 30 m (98.0 ft) from 97–0 km/h (60–0 mph), and can produce 1.41 G while cornering. \newline \newline 
Title: Porsche 911 \newline \newline Porsche 911 GT1 on display at the Porsche Museum The Porsche 911 GT1 is a race car that was developed in 1996 for the GT1 class in the 24 Hours of Le Mans. In order to qualify for GT racing, 25 road-going models were built to achieve type homologation. The engine in the GT1 is rated at 608 PS (447 kW; 600 hp) (544 PS (400 kW; 537 hp) for the road version) and accelerated from 0–97 km/h in 3.3 seconds. The top speed stood at 330 km/h (205 mph). (...) &

\textbf{GPT: } The Lamborghini Veneno is a limited production high performance sports car created to commemorate Lamborghini's 50th anniversary. It is based on the Lamborghini Aventador and was first showcased at the 2013 Geneva Motor Show with a price of \$4 million, making it one of the most expensive cars in the world. It has a top speed of 356 km/h, can accelerate from 0-97 km/h in 2.8 seconds, and can brake from 97-0 km/h in 30 meters while cornering at 1.41 Gs. The Porsche 911 GT1 is a race car developed in 1996 for the GT1 class in the 24 Hours of Le Mans. In order to qualify for GT racing, 25 road-going models were built to achieve type homologation. The engine produces 608 PS (600 hp) for the race version and 544 PS (537 hp) for the road version. It can accelerate from 0-97 km/h in 3.3 seconds and has a top speed of 330 km/h. (...) \newline
\textit{GPT: X -> X, Claude: X -> O} \newline \newline
\textbf{Claude: } Based on the background documents, it seems that Lamborghini Veneno would likely win a race against a Porsche 911. The Veneno has a higher top speed of 356 km/h compared to 330 km/h for the Porsche 911 GT1. The Veneno also has faster acceleration, able to go from 0 to 97 km/h in just 2.8 seconds compared to 3.3 seconds for the Porsche. Although the Porsche 911 GT1 was designed primarily as a race car and the Veneno was more of a limited production sportscar, the Veneno still has superior performance specifications. Given its faster speed, acceleration and handling, the Lamborghini Veneno model would probably win a race against a Porsche 911. \newline
\textit{GPT: X -> O, Claude: X -> O}
\\
\cline{2-3}
\hline
\end{tabular}
\caption{\textbf{An example of a gold context and paraphrased contexts in StrategyQA.} Below the paraphrased context are the results of each reader. \textit{GPT: X -> O} indicates that GPT reader got wrong with original context, but right with paraphrased context.}
\label{tab:stratqa}
\end{table*}

\begin{table*}
\begin{tabular}{c|p{7cm}p{7cm}}
\hline
Benchmark & Gold & Paraphrased \\
\hline
QASC & 
\textbf{Question: What are pollens? (A) eukaryotes (B) unicellular organisms \textcolor{red}{(C) building blocks of organisms} (D) antigens that cause allergy (E) bees (F) microscopic organisms (G) nucleotides composing polymers (H) diploid \newline Answer: (D)} \newline \newline
Any antigen that causes an allergy is called an allergen. Many of the common allergens are pollens and dusts.. & 
\textbf{GPT: }Allergens are substances that cause allergies, and they include common sources such as pollens and dusts. Pollens are not unicellular organisms or bees, but rather they are \textbf{\textcolor{red}{building blocks of organisms}} composed of nucleotides that form polymers. They are not diploid or microscopic, but they are known to be antigens that can trigger allergic reactions. \newline

\textit{GPT: O -> X, Claude: X -> X} \newline \newline
\textbf{Claude: } Here is a paraphrased summary of the background documents:\newline\newline Pollens are microscopic cells that contain the male gametes of plants and facilitate reproduction. They are produced by the anthers of stamens in flowering plants. When pollen grains get dispersed by wind, water, or insects and come in contact with the stigma of a flower, they germinate and produce a pollen tube to transfer the sperm cells to the ovule. \newline\newline Though pollens are important for plant \newline
\textit{GPT: O -> X, Claude: X -> X}
\\
\cline{2-3}
& \textbf{Question: Climate is generally described in terms of what? (A) sand (B) occurs over a wide range (C) forests (D) Global warming (E) rapid changes occur (F) local weather conditions (G) measure of motion (H) city life \newline Answer: (F)} \newline \newline
 Climate is generally described in terms of temperature and moisture.\newline\newline Fire behavior is driven by local weather conditions such as winds, temperature and moisture. & 
\textbf{GPT: } The climate is typically defined by its temperature and moisture levels. The behavior of fires is influenced by local weather conditions, which may include wind, temperature, and moisture. \newline
\textit{GPT: X -> O, Claude: X -> O} \newline \newline
\textbf{Claude: } Based on the background documents, climate is generally described in terms of temperature and moisture or local weather conditions, not the other options. \newline
\textit{GPT: X -> O, Claude: X -> O}
\\
\hline
\end{tabular}
\caption{\textbf{Examples of gold contexts and paraphrased contexts in QASC.} Below the paraphrased context are the results of each reader. \textit{GPT: X -> O} indicates that GPT reader got wrong with original context but right with paraphrased context. The first example shows a failure case, where the distractor option misleads the paraphraser to generate irrelevant information (see the \textbf{\textcolor{red}{red-colored}} part). The second example, on the other hand, shows a success case.}
\label{tab:qasc}
\end{table*}

\end{document}